%% file: acl_latex.tex
\pdfoutput=1

\documentclass[11pt]{article}
\usepackage{adjustbox}  
\usepackage[final]{acl}
\usepackage{makecell} 
\usepackage{pifont}

\usepackage{times}
\usepackage{latexsym}
\usepackage{amsmath}
\usepackage{float}
\usepackage{textcomp}
\usepackage[T1]{fontenc}

\usepackage[utf8]{inputenc}

\usepackage{microtype}

\usepackage{inconsolata}

\usepackage{graphicx}
\usepackage{booktabs}
\usepackage{multirow}
\usepackage{enumitem}

\title{Are Optimal Algorithms Still Optimal? Rethinking Sorting in LLM-Based Pairwise Ranking with Batching and Caching}

\author{%
  \normalfont
  Juan Wisznia\textsuperscript{1,3}, Cecilia Bolaños\textsuperscript{1,2},\\[-0.2em]
  Juan Tollo\textsuperscript{1}, Giovanni Marraffini\textsuperscript{1,3}, Agustín Gianolini\textsuperscript{1},\\[-0.2em]
  Noe Hsueh\textsuperscript{1}, Luciano Del Corro\textsuperscript{1,3}\\[0.6em]
  \small
  \{jwisznia, cbolanos, jtollo, agianolini, nhsueh, ldelcorro\}@dc.uba.ar, giovanni.marraffini@gmail.com\\[0.7em]
  \textsuperscript{1}Departamento de Computación, FCEyN, Universidad de Buenos Aires\\
  \textsuperscript{2}Instituto de Ciencias de la Computación, FCEyN, Universidad de Buenos Aires\\
  \textsuperscript{3}Lumina Labs\textsuperscript{*}
}

\begin{document}
\maketitle
\renewcommand{\thefootnote}{\fnsymbol{footnote}}
\footnotetext[1]{Work initiated at Lumina Labs.}
\renewcommand{\thefootnote}{\arabic{footnote}}


\begin{abstract} We introduce a novel framework for analyzing sorting algorithms in pairwise ranking prompting (PRP), re-centering the cost model around LLM inferences rather than traditional pairwise comparisons. While classical metrics based on comparison counts have traditionally been used to gauge efficiency, our analysis reveals that expensive LLM inferences overturn these predictions; accordingly, our framework encourages strategies such as batching and caching to mitigate inference costs. We show that algorithms optimal in the classical setting can lose efficiency when LLM inferences dominate the cost under certain optimizations. \end{abstract}

\section{Introduction}

LLMs have ushered in a new era of language understanding \cite{brown2020languagemodelsfewshotlearners}. Alongside these developments, LLM-based reranking has emerged in the information retrieval (IR) domain \cite{nogueira2020documentrankingpretrainedsequencetosequence, rankt5, ma2023zeroshotlistwisedocumentreranking, sun2024chatgptgoodsearchinvestigating}. Instead of using custom fine-tuned rankers, off-the-shelf LLMs—often combined with a first-stage retriever can refine search results in a zero-shot manner. The practical significance of reranking is evident in its rapid commercial adoption, with major cloud platforms now offering it as a core functionality. LLM-based reranking enables robust ranking quality without the overhead of dataset-specific models, which is crucial, for example, for the widespread adoption of Retrieval-Augmented Generation across both cloud-based and on-prem deployments.

A notable exemplar in zero-shot LLM-based reranking is Pairwise Ranking Prompting (PRP) \cite{qin-etal-2024-large, luo-etal-2024-prp}, which compares two candidate documents. Despite its conceptual elegance and model-agnostic nature, PRP faces significant computational challenges; in practice—each pairwise comparison requires an expensive LLM inference, making a naive all-pairs approach prohibitively costly \cite{qin-etal-2024-large}. This has prompted both researchers and practitioners to adopt classical sorting algorithms for minimizing the number of comparisons \cite{qin-etal-2024-large} as they offer theoretical guarantees. 

We argue that classical analysis is not adequate for PRP as it treats each comparison as an atomic, uniform-cost operation, whereas in an LLM-based system, each comparison is an expensive inference call. This gap between classical and LLM-centric views can invert conventional wisdom under certain basic optimizations, causing algorithms that appear optimal under traditional assumptions to underperform in real-world scenarios, and vice versa.

To address these limitations, we introduce a framework that redefines how ranking algorithms are analyzed in an LLM context. Rather than merely counting comparisons, we focus on LLM inference calls as the primary cost driver. We show that basic optimizations—such as caching and batch inference—can significantly alter algorithms' efficiency. Furthermore, we propose Quicksort as an efficient reranking algorithm, demonstrating its potential when leveraging these optimizations. To the best of our knowledge, this is the first time Quicksort has been applied in this context.

Caching and Batching have no effect on algorithm ranking performance; the exact same comparisons will be performed but much faster. While caching repeated queries and batching independent operations are seemingly trivial adaptations, they significantly affect the choice of the optimal algorithm challenging previous results \cite{qin-etal-2024-large, Zhuang_2024}. For instance, Heapsort is no longer the preferred choice. A mere batch size of 2 will result in Quicksort generating 44\% less inference calls compared to Heapsort.

We validate our findings on standard ranking benchmarks (TREC DL 2019 and 2020 \cite{craswell2020overviewtrec2019deep, craswell2021overviewtrec2020deep} and BEIR \cite{thakur2021beirheterogenousbenchmarkzeroshot}). By re-framing sorting theory around real-world LLM inference costs, we offer both practical guidance for zero-shot reranking and a theoretical basis for understanding algorithmic efficiency under modern IR constraints. 

\section{Related Work}
Traditional IR systems require extensive labeled data and struggle with cross-domain generalization \cite{Matveeva2006HighAccuracy, Wang2011Cascade}. LLMs have transformed this landscape by enabling zero-shot ranking. PRP emerged then as a particularly effective technique \cite{qin-etal-2024-large, luo-etal-2024-prp}. PRP's key advantage lies in its model-agnostic nature- by comparing document pairs through simple prompts, it can leverage any LLM without training or access to model internals, making it especially valuable as newer models emerge. However, PRP faces significant computational challenges as each pairwise comparison requires an expensive LLM inference, with costs scaling quadratically with document count.

To address these computational demands, recent work has incorporated sorting algorithms into the PRP framework \cite{qin-etal-2024-large, Zhuang_2024}. 
While theoretically well-grounded, these approaches adopt the cost framework of traditional sorting theory, where comparisons are treated as atomic operations with uniform costs. However, in LLM-based ranking, inferences are orders of magnitude more expensive than other operations. This mismatch between classical cost assumptions and LLM-specific characteristics suggests the need to reevaluate sorting algorithm selection and optimization for real-world performance. 

\section{Revisiting sorting algorithms}
\label{subsec:revisiting_sorting_algorithms}

This section examines how small yet impactful opimizations (caching, batching, and top-k extraction) in the context of classical algorithms—Bubblesort, Quicksort, and Heapsort can significantly shift which algorithm is most efficient in LLM-based ranking. While these adaptations are not exhaustive, they demonstrate how our framework redefines efficiency based on LLM-specific costs, where reducing inference steps matters more than traditional complexity metrics. Importantly, these optimizations preserve the final ranking outcome: the same comparisons are performed but are batched or reused, leading to fewer inference calls and a much faster process. Table \ref{tab:algorithms} summarizes the optimizations applicable to each algorithm.

\begin{table}[h]
    \small
    \centering
    \begin{tabular}{lccc}
        \toprule
        \textbf{Algorithm} & \textbf{Batching} & \textbf{Caching} & \textbf{Top-k Efficiency} \\
        \midrule
        Heapsort   & \ding{55} & \ding{55} & \ding{51} \\
        Bubblesort & \ding{55} & \ding{51} & \ding{51} \\
        Quicksort  & \ding{51} & \ding{55} & \ding{51}\footnotemark \\
        \bottomrule
    \end{tabular}
    \caption{Summary of optimization techniques under LLM-centric costs.}
    \label{tab:algorithms}
\end{table}
\footnotetext{Using Partial Quicksort \cite{martinez2004partial}.}

\begin{figure}[htb]
  \hfill  
  \includegraphics[width=0.90\linewidth]{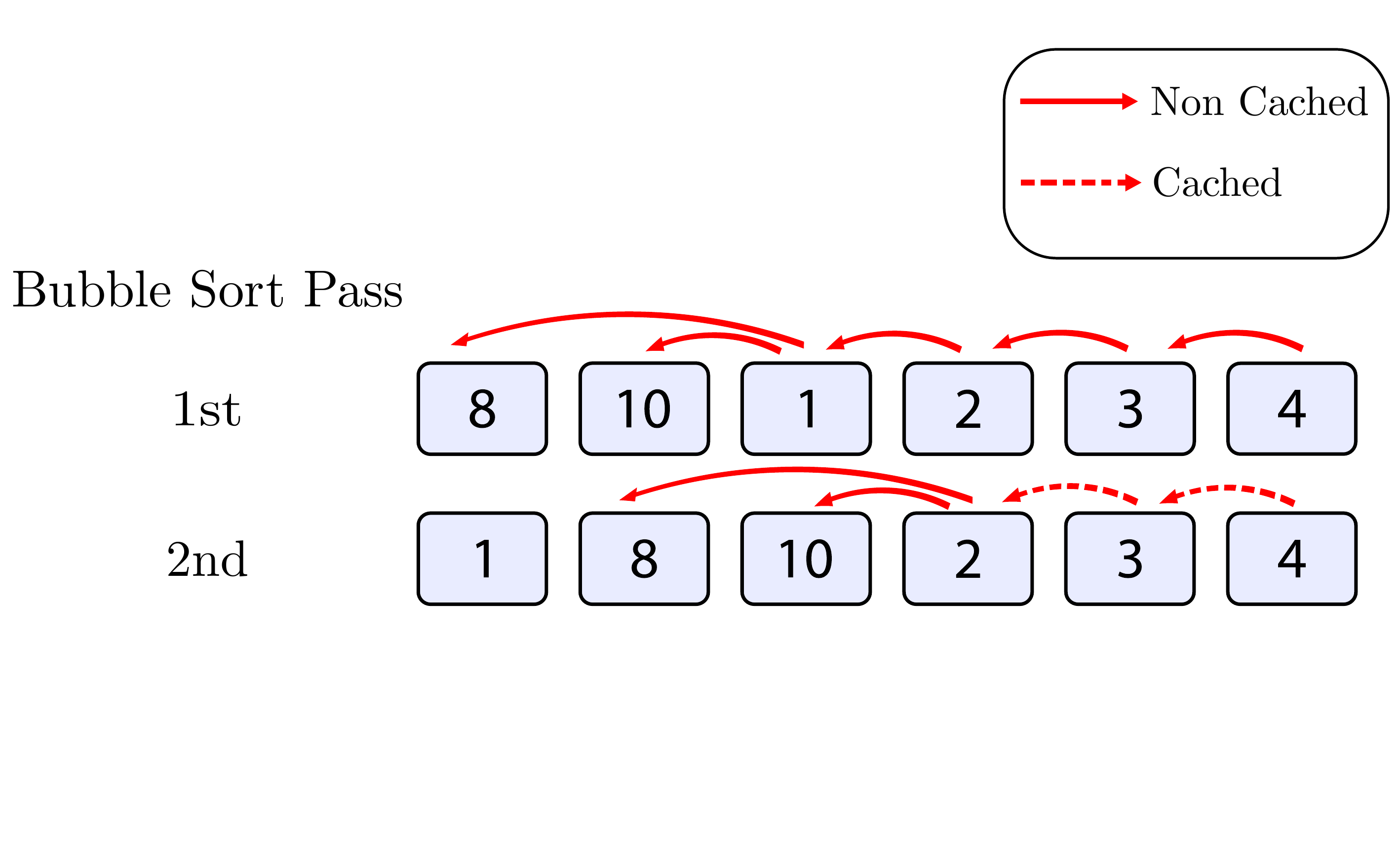}
  \vspace{-3em}
  \caption{Bubblesort with Caching. Solid arrows show inferences, dashed arrows cached comparisons.}
  \label{fig:Bubblesort}
\end{figure}

\noindent\textbf{Heapsort} has been favored in early PRP research \citet{qin-etal-2024-large} for its $O(n \log n)$ complexity and natural support for top-k extraction. However, it cannot be adapted to batching or caching due to its binary tree structure. Each comparison is inherently sequential and unique. This makes it impossible to group comparisons into a single inference step (batching) or to reuse prior results (caching) effectively.

\begin{figure*}[h]
  \includegraphics[width=0.48\linewidth]{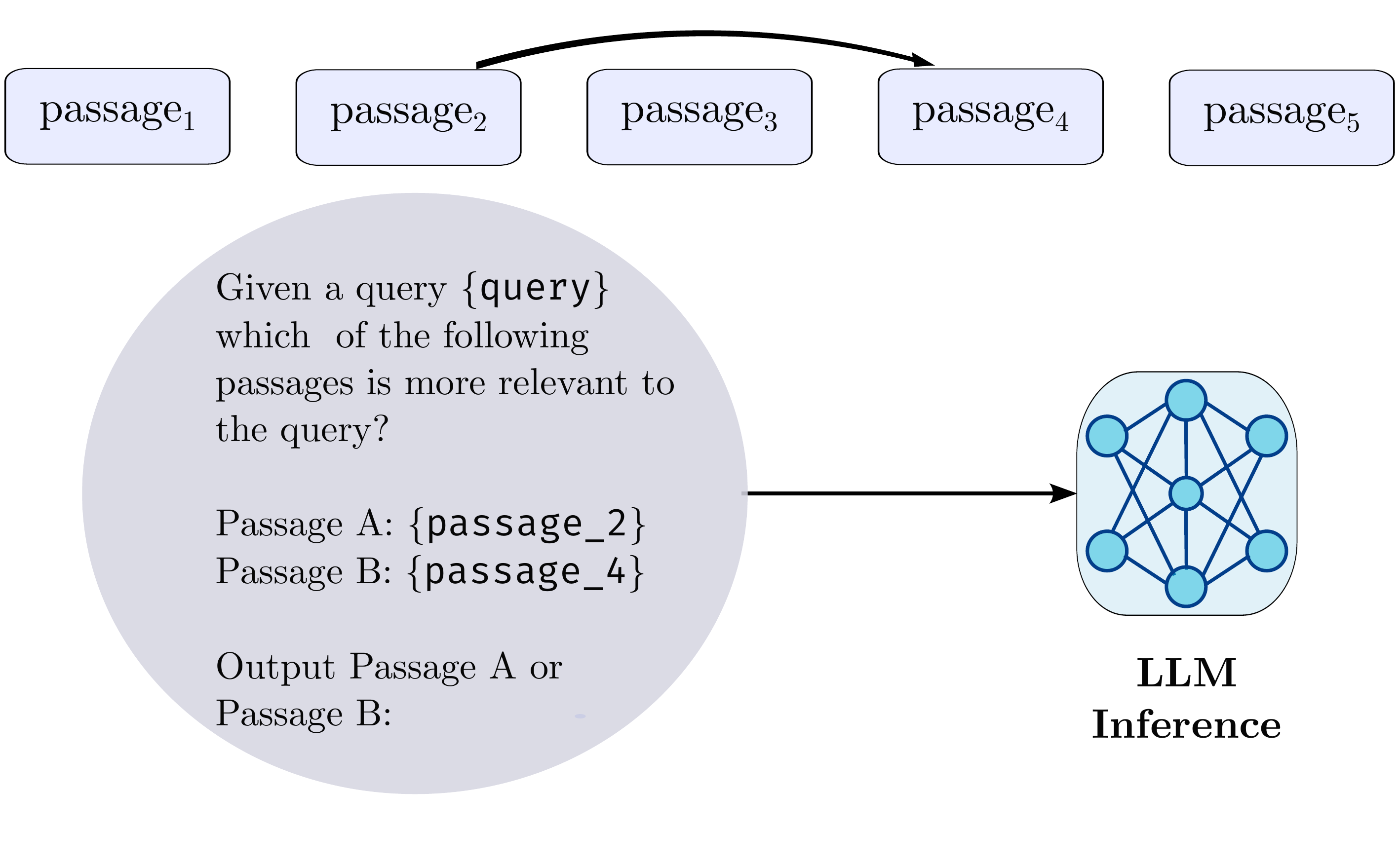} \hfill
  \includegraphics[width=0.48\linewidth]{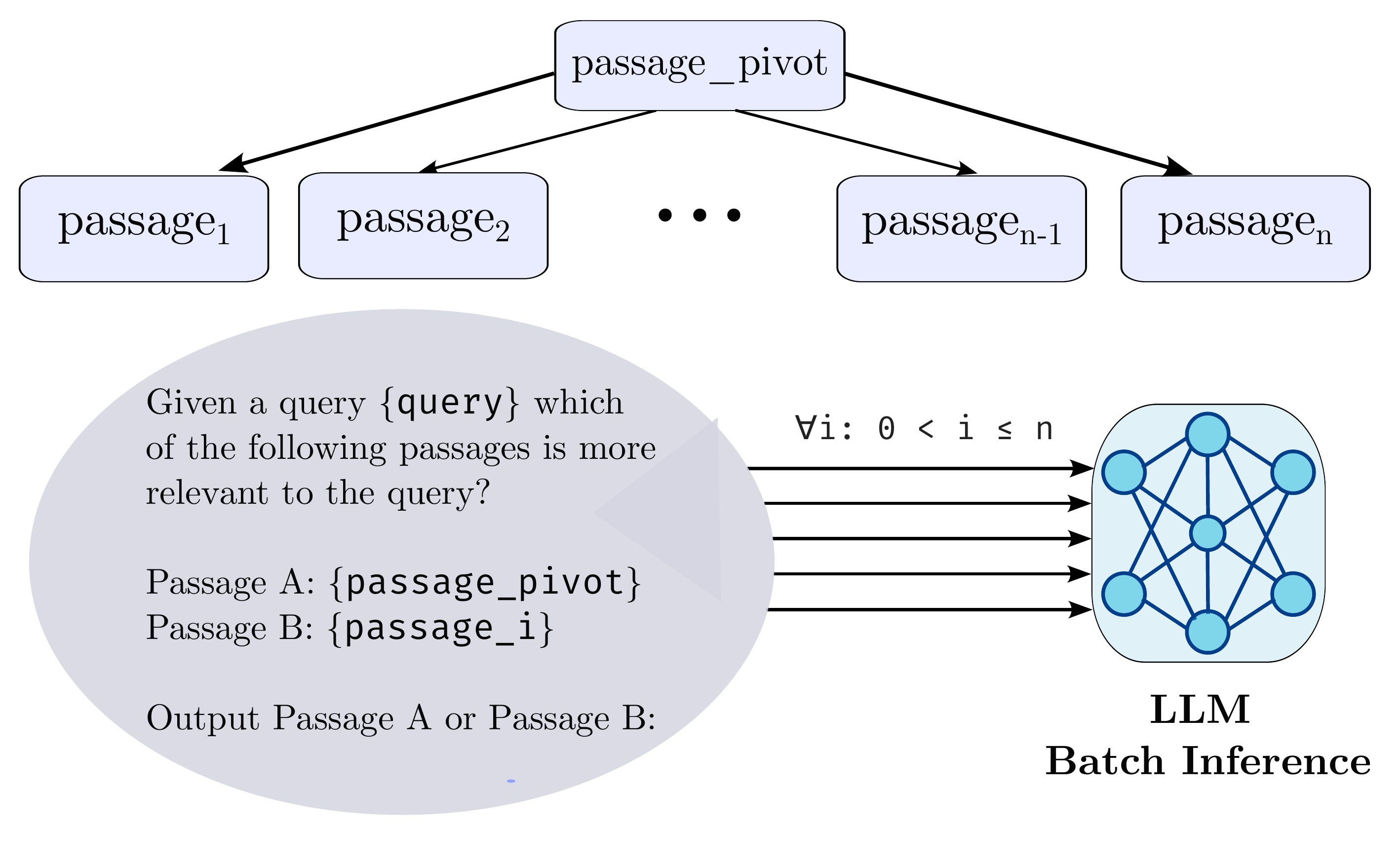}
  \caption{a. One comparison per inference as classic analysis for these algorithms. b. Multiple comparisons per inference making gaining more information per inference.}
  \label{fig:batch_quick}
\end{figure*}

\noindent\textbf{Bubblesort} has been considered expensive due to its 
$O(n^2)$ complexity, but this can be adapted via caching from its repeated adjacent comparisons across passes (Figure~\ref{fig:Bubblesort}). The memory overhead remains negligible, requiring only a small dictionary to store prior results. While its pairwise swap structure precludes batching (comparisons cannot be grouped into single inferences), it inherently supports top-k extraction \citep{qin-etal-2024-large}, enhancing its practicality for ranking applications.

\noindent\textbf{Quicksort} uniquely enables batching through its partition phase, where multiple elements can be evaluated simultaneously against a pivot (Figure~\ref{fig:batch_quick}). However, it has limited potential for caching, as pivot comparisons are typically non-repeating. Despite this, the Partial Quicksort variant \citep{martinez2004partial} enhances its efficiency by enabling early termination for top-k extraction. To the best of our knowledge, we are the first to introduce Quicksort in PPR as prior research focused on Heapsort and Bubblesort due to their top-k properties.

\section{Experimental Setup} 

\noindent\textbf{Hardware:} Our analysis is theoretical and agnostic to hardware. However, to validate that our cost assumptions align with practical throughput behavior, we ran two lightweight empirical checks on NVIDIA A100 (40GB), RTX 3090, and RTX 2080 Ti. These include single forward-pass latency measurements across batch sizes and GPUs, and full PRP reranking with Quicksort and Heapsort at batch sizes 2 and 128 for the A100 (see Section~\ref{sec:results_discussion}).

\noindent\textbf{Metric:} Instead of focusing on traditional comparison counts, we shifted to the number of LLM inference calls, which are the dominant computational cost. Each inference—regardless of token count or monetary cost—is treated as a uniform cost unit. We disregard token counts and dollar costs because these are determined by the dataset and pre-trained model. Moreover, standard pre-processing (e.g., chunking/truncation) ensures uniformity across documents. We show mean and standard deviation across datasets and LLMs. Individual results can be found in Appendix~\ref{sec:appendix}.

\noindent\textbf{LLMs:} Following \citet{qin-etal-2024-large, Zhuang_2024} we used: Flan-T5-L (780M), Flan-T5-XL (3B), Flan-T5-XXL (11B) \cite{chung2022scalinginstructionfinetunedlanguagemodels}, Mistral-Instruct (7B) \cite{jiang2023mistral7b}, and Llama-3-Instruct (8B) \cite{dubey2024llama3herdmodels}. For the latency analysis we implemented batch processing with Flan-T5-Large using the Hugging Face Transformers library \cite{wolf2020huggingfacestransformersstateoftheartnatural}.

\noindent\textbf{Algorithms:} (1) Bubblesort, (2) Quicksort with median-of-three pivot strategy (other strategies are shown in Appendix~\ref{sec:appendix}), and (3) Heapsort.

\noindent\textbf{Datasets:} TREC DL 2019 (43 queries) and 2020 (200) \cite{craswell2020overviewtrec2019deep, craswell2021overviewtrec2020deep} as well as subsets from BEIR \cite{thakur2021beirheterogenousbenchmarkzeroshot}: Webis-Touche2020 (49), NFCorpus (295), Large-Scifact (300), TREC-COVID (50), FiQA (648), and DBpedia-Entity (400). Following standard practices, we re-ranked the top 100 BM25-retrieved documents per query \cite{Robertson2009ThePR, qin-etal-2024-large, Zhuang_2024, luo-etal-2024-prp} to identify the top-10 most relevant ones efficiently.

\section{Results and Discussion}
\label{sec:results_discussion}

\begin{figure*}[t]
    \centering
    \vspace{-1.5cm} 
    \resizebox{\linewidth}{!}{\includegraphics{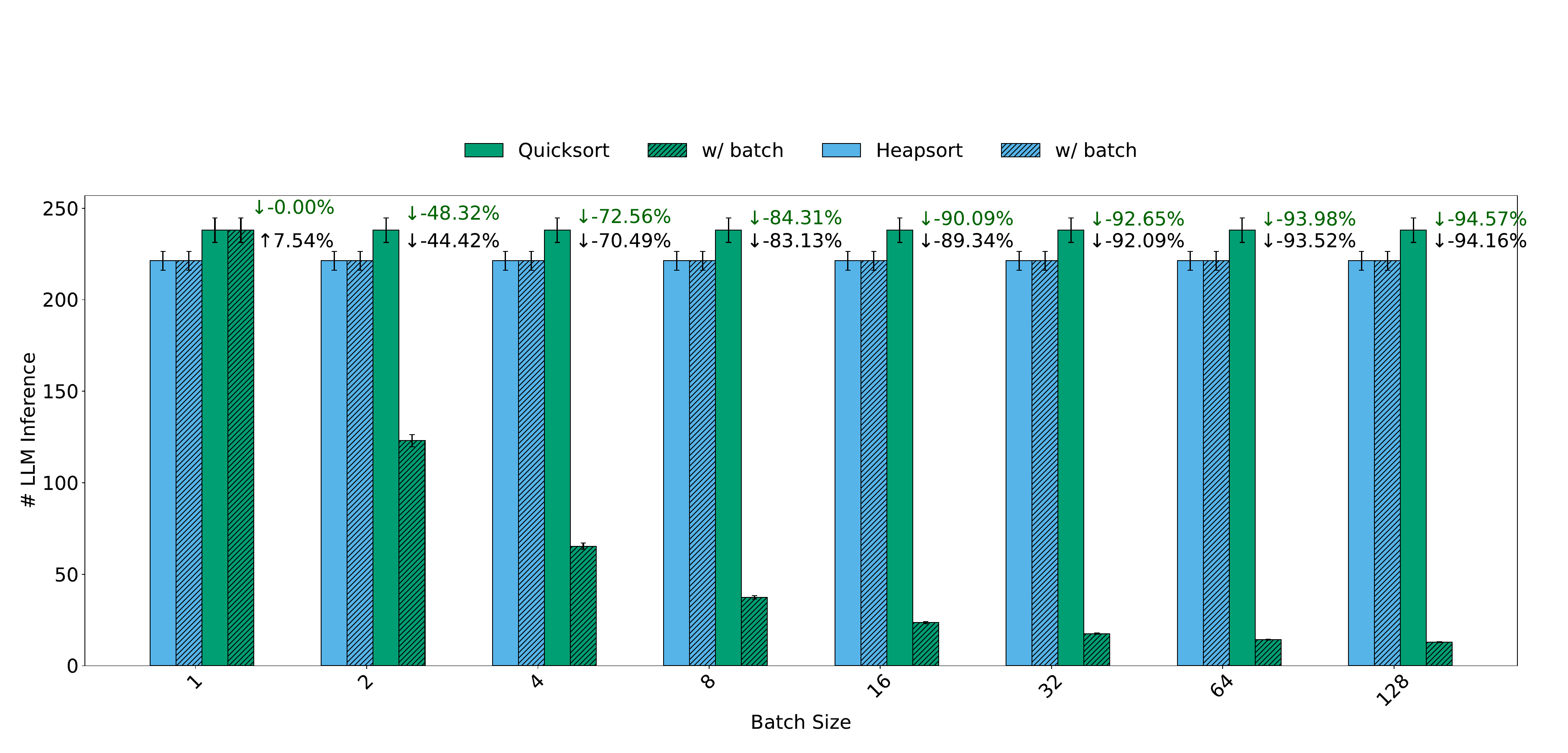}}
    \caption{Mean and SD inference count for Quicksort and Heapsort across batch sizes. Black number: Heapsort vs. Quicksort using batching gain; Green number: Quicksort batching vs. no batching gain.}
    \label{fig:inferences}
\end{figure*}

\noindent\textbf{Cost model analysis:} Figure~\ref{fig:inferences} illustrates the number of inferences performed by Heapsort and Quicksort across different batch sizes. When the batch size is set to 1 (equivalent to counting individual comparisons), Heapsort emerges as the most efficient algorithm consistent with traditional sorting analysis and previous results \cite{qin-etal-2024-large, Zhuang_2024}. However, as the batch size increases, Quicksort is able to significantly outperform as multiple comparisons can be run in parallel. For instance, with a batch size of 2, the average number of inference calls is reduced already by almost 45\%.

 \begin{figure}[h]
  \includegraphics[width=1.0\columnwidth]{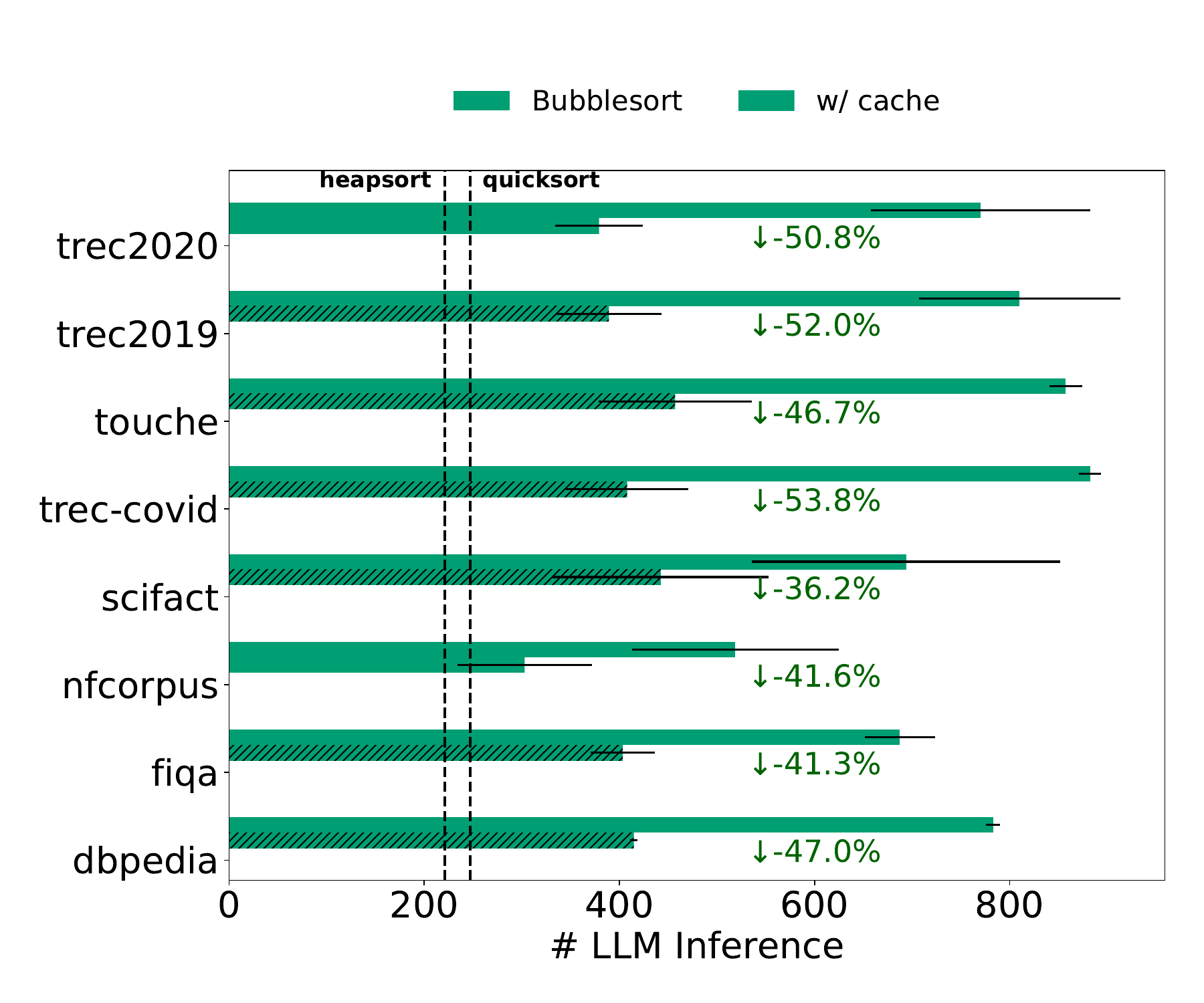}
   \label{'cache'}
    \caption{Mean and SD inference count for Bubblesort, with and without cache. Green numbers indicate the percentage gain with cache. The dashed line represents the mean inference count for Heapsort and Quicksort.}
   \label{fig:buble}
\end{figure}

Figure~\ref{fig:buble} compares the number of inferences performed by Bubblesort with and without cache. Bubblesort benefits significantly more from caching at a minimal storage overhead. This is because Bubblesort involves repeated comparisons, many of which can be cached, reducing the total number of inferences by an average of 46\%.

Importantly, despite these optimizations reducing the number of LLM inferences, they do not alter the final ranking outcome. The same comparisons are performed, but they are either batched together or retrieved from cache rather than recomputed, leading to fewer inference calls and a much faster process.

\begin{figure}[h]
    \centering
    \includegraphics[width=\linewidth]{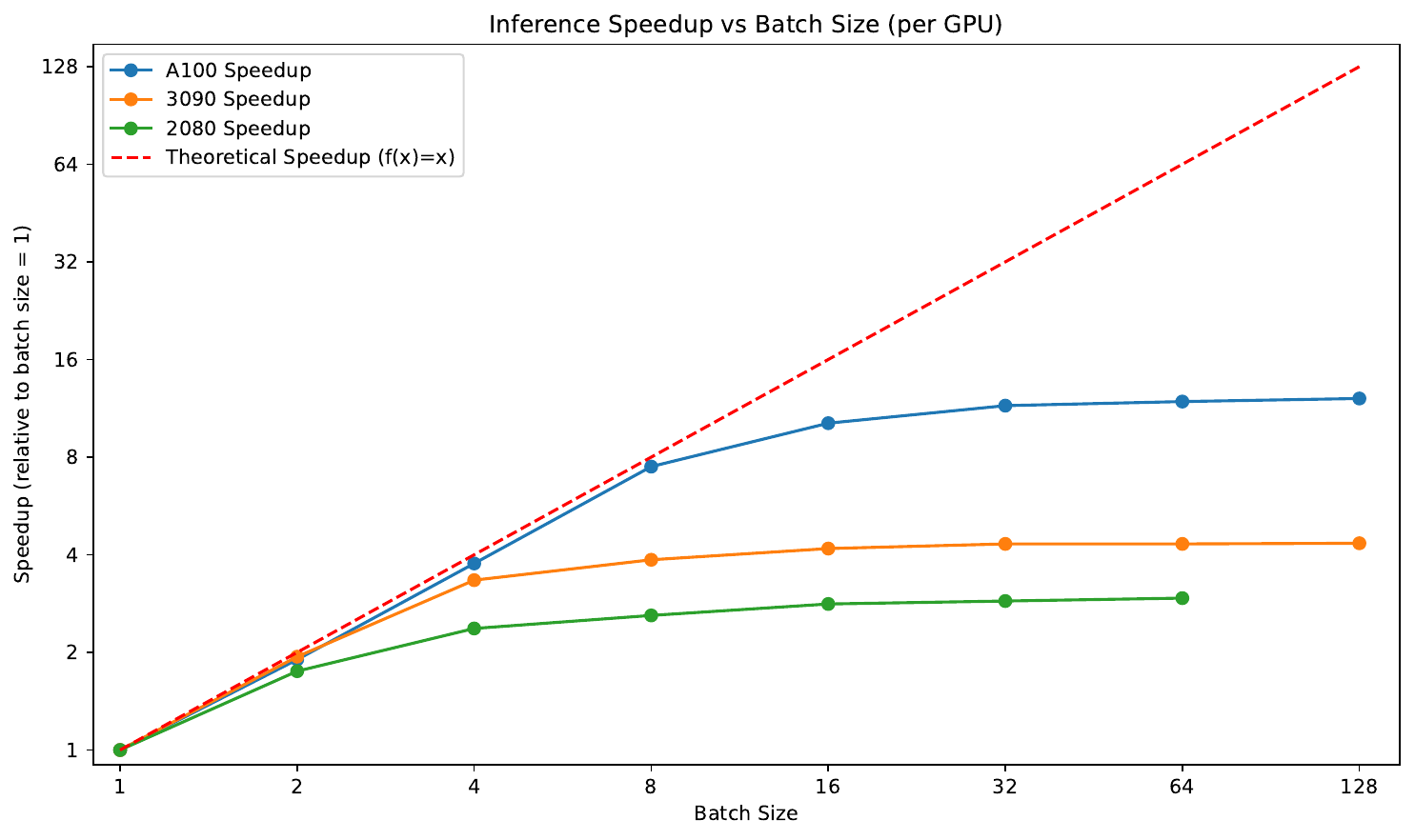}
    \caption{Speed-up vs batch size for Flan-T5-Large (log–log). Dashed red: ideal linear scaling.}
    \label{fig:latency_gpu_speedup}
\end{figure}

\vspace{0.4em}
\noindent\textbf{Latency Analysis:} 
Figure~\ref{fig:latency_gpu_speedup} shows single-pass speed-ups on A100, RTX 3090, and RTX 2080 Ti for different batch sizes. A100 achieves near-ideal scaling up to batch size 8, with throughput continuing to improve—albeit with diminishing returns—up to batch 128. On 3090 and 2080 Ti, ideal scaling occurs up to batch sizes 2 and 4, respectively, with throughput saturating between batch sizes 32 and 64. These results indicate that while theoretical efficiency peaks at larger batch sizes, practical efficiency is constrained by GPU architecture. The point at which near-ideal conditions are met before saturation sets in is GPU-dependent.

We also ran the full PRP pipeline over BEIR on the A100 using both batch size 2 and 128 for Quicksort and Heapsort. At batch 128, Quicksort is \(5.52\times\) faster than Heapsort while achieving similar nDCG@10 (See Appendix~\ref{sec:appendix}, Tables~\ref{tab:dataset-comparison-1}--\ref{tab:dataset-comparison-2})
. Experiments show that the theoretical gains from batching and algorithmic design hold in end-to-end ranking performance.

\begin{figure}[h]
\includegraphics[width=1\columnwidth]{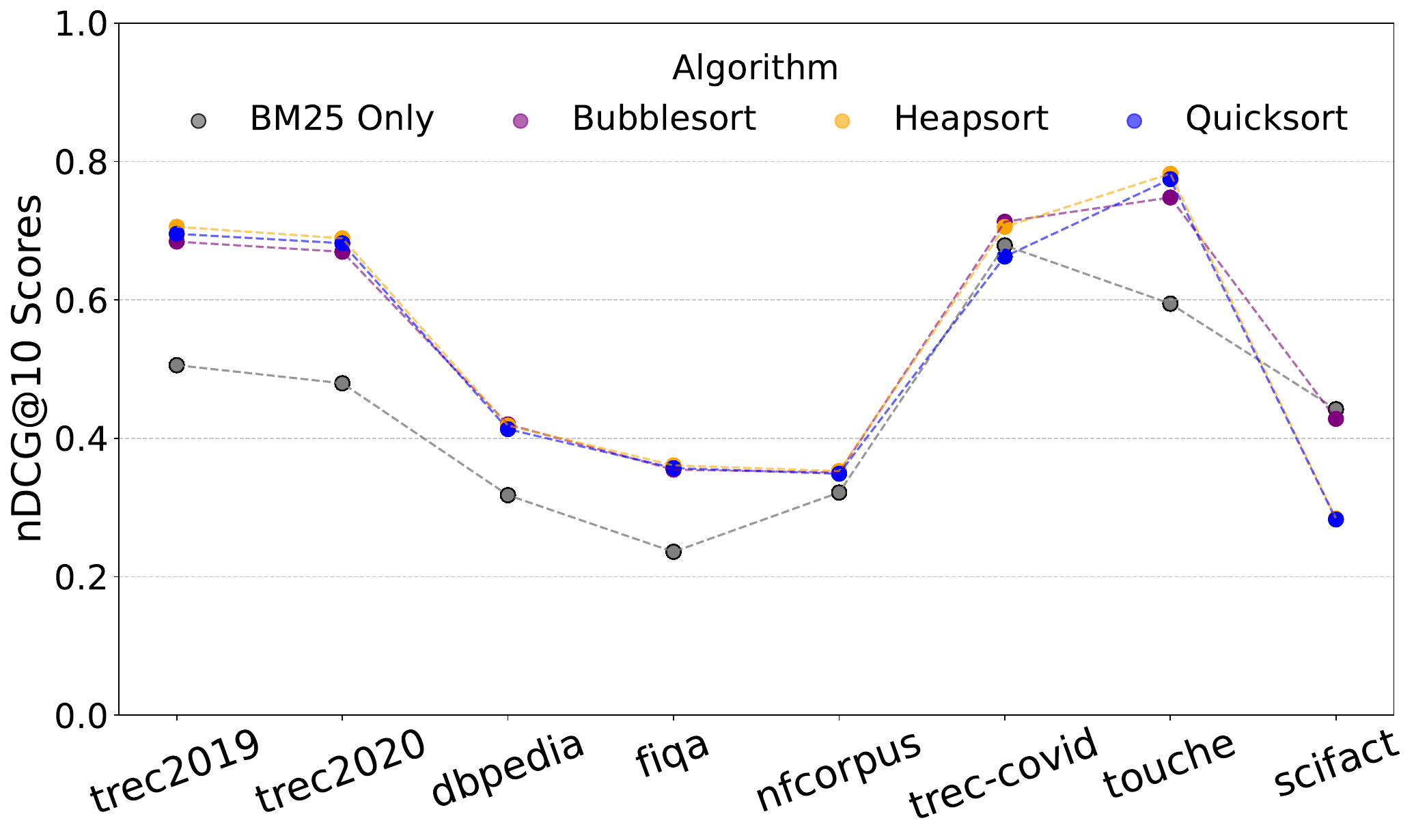}
\label{'ndcg'}
\caption{Algorithms' performance across datasets.}
\label{fig:ndcg}
\end{figure}

\noindent\textbf{Ranking Performance:} Figure~\ref{fig:ndcg} shows that the ranking performance of all these algorithms across optimization settings remains relatively stable for a given dataset, allowing users to prioritize computational efficiency and hardware constraints before performance when choosing an algorithm.

Findings provide a detailed insight of sorting algorithms behavior in LLM-based pairwise ranking, highlighting their respective benefits and drawbacks, enabling users to select the most suitable algorithm based on their specific resources and requirements. More specifically:

\noindent\textbf{Quicksort} is ideal for latency-sensitive applications with batch sizes $\geq 2$, leveraging hardware parallelism to outperform alternatives. 

\noindent\textbf{Bubblesort} achieves a susbtantial efficiency gain with caching. It's remarkable performance in some datasets like scifact and touche2020 makes it a more competitive choice with the new adaptation. Bubblesort tends to be effective in the context of LLMs in which pairwise transitivity is not guarantied. Pairwise adjacent comparisons seemed to be more stable and bring better results in the context of PRP \cite{luo-etal-2024-prp}.

\noindent\textbf{Heapsort}, once the gold standard for its theoretical logarithmic complexity, its advantage emerges only with no batching (rarely seen in LLM ranking).

\section{Conclusion}

We introduced a framework for optimizing sorting algorithms in LLM-based pairwise ranking by prioritizing inference calls over comparison counts. We found that classical efficiency assumptions break down under LLM workloads, revealing Quicksort as a natural, yet unexplored, choice of algorithm. This demonstrates that inference efficiency is a property deeply tied to algorithmic design. We hope this framework encourages further exploration of algorithms better aligned with LLM cost structures.

\section{Limitations}
While our work showcases the efficacy of batching and caching optimizations in mitigating the high inference costs of LLM-based pairwise ranking, certain limitations remain. First, sorting algorithms work best when transitivity in pairwise comparisons holds, but LLMs can yield inconsistent judgments for near-equivalent or context-sensitive documents. Addressing this inconsistency requires dedicated methods to detect and resolve intransitive preferences, which remains an open area of research. Future work could examine how much performance is degraded and whether ranking algorithms that do not assume transitivity can actually offer any practical advantage.

Additionally, although our experiments were limited to medium-sized LLMs for budgetary and computational reasons, larger models could further amplify the benefits observed here. Future research should explore how our framework performs with these more powerful models, potentially unlocking even greater gains in inference efficiency. Moreover, hybrid methods that unify the strengths of multiple algorithms, as well as active ranking strategies or noisy sorting algorithms \cite{ASAP2020, bai2023sortingpredictions}, are fully compatible with our approach: they rely on additional computations separate from the LLM inferences themselves, thereby enabling more informed—and thus fewer—LLM queries. Ultimately, our findings underscore the need for ongoing algorithmic innovation that exploits LLM-specific cost structures, paving the way for more efficient, scalable, and broadly applicable ranking solutions.

\bibliography{custom}

\clearpage
\onecolumn

\appendix

\section{Appendix}

\label{sec:appendix}

In this appendix, we present a comparison of different methods across the \textbf{BEIR} and \textbf{TREC} datasets. Each table follows the same structure and reports NDCG@10 (Normalized Discounted Cumulative Gain), the number of inferences, and the number of comparisons (\#Inferences and \#Comparisons) for each method. Also, latency (in seconds) is reported for Quicksort and Heapsort over the BEIR suite and corresponds to end-to-end PRP execution over each dataset on an A100 40GB GPU. Quicksort is analyzed using four pivot selection strategies: the original Hoare's method, the middle-element selection, random and median-of-three strategies \cite{Hoare62a, sedgewick75}.

To highlight performance differences, we emphasize the best-performing algorithm for each dataset and LLM model in \textbf{black}, while the second-best is \underline{underlined}. Additionally, the tables distinguish between two computational scenarios: (1) \texttt{cached}: The number of inferences made to the LLM.
(2) \texttt{non-cached}: The number of comparisons performed using precomputed results, avoiding additional inferences.
All results are presented with a batch size of 2 and 128 to show batch inference efficiency.

\begingroup
\input{table_beir}
\input{table_trec}
\endgroup

\end{document}

%% file: table_beir.tex
\begin{table*}[!ht]
\small
\centering
\setlength{\tabcolsep}{1.5pt}
\begin{adjustbox}{width=\textwidth,keepaspectratio}
\begin{tabular}{@{}ll*{12}{r}@{}}
\toprule
&  & \multicolumn{4}{c}{dbpedia} & \multicolumn{4}{c}{nfcorpus} & \multicolumn{4}{c}{fiqa} \\
\cmidrule(lr){3-6} \cmidrule(lr){7-10} \cmidrule(l){11-14}
\# & Methods & NDCG@10 & \#comp & \#inf & Lat. & NDCG@10 & \#comp & \#inf & Lat. & NDCG@10 & \#comp & \#inf & Lat. \\
\midrule
 & BM25 & 0.318 & -- & -- & -- & 0.322 & -- & -- & -- & 0.240 & -- & -- & -- \\
\midrule
\multirow{11}{*}{\rotatebox[origin=c]{90}{Flan-t5-large}}  
 & heapsort  \#
   & 0.413 & 225.1 & 225.1 & 24.8 
   & 0.335 & 160.0 & 160.0 & 18.4 
   & \textbf{0.313} & 200.3 & 200.3 & 26.1 \\
 & quicksort (original, b=2) & 0.403 & 245.3 & 126.8 &34.1 & 0.321 & 171.0 & 88.9 & 16.8 & 0.284 & 253.3 & 130.8 &27.7 \\

 & quicksort (original, b=128) & 0.403 & 245.3 & 13.2 & 4.7 & 0.321 & 171.0 & 11.1& 3.0  & 0.284 & 253.3 & 13.6 & 4.8 \\

 & quicksort (random, b=2) & 0.414 & 236.9 & 122.4 & -- & 0.322 & 181.3 & 94.1& --  & 0.282 & 246.0 & 127.0& --  \\

 & quicksort (random, b=128) & 0.405 & 241.5 & 13.0 & -- & 0.322 & 171.6 & 10.9& --  & 0.289 & 246.6 & 13.2 & -- \\

 & quicksort (middle, b=2) & 0.410 & 231.9 & 119.9 & --  & 0.315 & 168.9 & 87.8 & -- & 0.277 & 243.2 & 125.6 & -- \\

 & quicksort (middle, b=128) & 0.410 & 231.9 & 12.8 & -- & 0.315 & 168.9 & 10.9 & --  & 0.277 & 243.2 & 13.2 & -- \\

 & quicksort (median of three, b=2) & 0.414 & 255.2 & 115.6 & -- & 0.326 & 187.0 & 83.7 & -- & \underline{0.295} & 284.9 & 128.7 & -- \\

 & quicksort (median of three, b=128) & 0.414 & 255.2 & 12.3 & -- & 0.326 & 187.0 & 10.5 & --  & 0.295 & 284.9 & 13.1 & -- \\ & bubblesort (classic) 
   & \underline{0.415} & 777.6 & 777.6 & -- 
   & \underline{0.343} & 593.9 & 593.9 & -- 
   & 0.295 & 662.1 & 662.1 & -- \\
 & bubblesort (cached) 
   & \textbf{0.415} & 777.6 & 360.4 & -- 
   & \textbf{0.343} & 593.9 & 242.2 & -- 
   & 0.295 & 662.1 & 235.3 & -- \\
\midrule
\multirow{11}{*}{\rotatebox[origin=c]{90}{Flan-t5-xl}} 
 & heapsort 
   & 0.419 & 229.3 & 229.3 & -- 
   & {\color{black}0.353} & 144.9 & 144.9 & -- 
   & \textbf{0.361} & 224.5 & 224.5 & -- \\
 & quicksort (original, b=2) & 0.404 & 238.6 & 123.3 & -- & 0.345 & 160.6 & 83.7 & --  & 0.338 & 209.1 & 108.3 & -- \\

 & quicksort (original, b=128) & 0.404 & 238.6 & 12.7 & -- & 0.345 & 160.6 & 10.9 & -- & 0.338 & 209.1 & 11.9 & -- \\

 & quicksort (random, b=2) & 0.412 & 221.6 & 114.7 & -- & 0.343 & 168.7 & 87.6& --  & 0.345 & 211.3 & 109.5 & -- \\

 & quicksort (random, b=128) & 0.411 & 230.8 & 12.4 & -- & 0.344 & 159.0 & 10.5& --  & 0.338 & 211.3 & 12.0 & -- \\

 & quicksort (middle, b=2) & 0.410 & 221.3 & 114.5 & -- & \underline{0.353} & 157.6 & 82.1 & --  & 0.341 & 205.4 & 106.5 & -- \\

 & quicksort (middle, b=128) & 0.410 & 221.3 & 12.2 & -- & \textbf{0.353} & 157.6 & 10.4 & -- & 0.341 & 205.4 & 11.9 & -- \\

 & quicksort (median of three, b=2) & 0.413 & 234.6 & 106.3 & -- & 0.349 & 184.9 & 82.8 & -- & 0.357 & 226.6 & 102.7 & -- \\

 & quicksort (median of three, b=128) & 0.413 & 234.6 & 11.6 & -- & 0.349 & 184.9 & 10.3 & -- &\underline{ 0.357} & 226.6 & 11.1 & -- \\
& bubblesort (classic) 
   & \underline{0.420} & 788.2 & 788.2 & -- 
   & 0.351 & 443.8 & 443.8  & --
   & 0.355 & 712.8 & 712.8 & --\\
 & bubblesort (cached) 
   & \textbf{0.420} & 788.1 & 376.0 & --
   & 0.351 & 443.8 & 189.6 & --
   & 0.355 & 712.8 & 332.5 & -- \\
\bottomrule
\end{tabular}
\end{adjustbox}
\caption{Comparison of different methods across \textbf{DBPedia}, \textbf{NFCorpus}, and \textbf{FiQA} datasets.}
\label{tab:dataset-comparison-1}
\end{table*}

\begin{table*}[!ht]
\small
\centering
\setlength{\tabcolsep}{1.5pt}
\begin{adjustbox}{width=\textwidth,keepaspectratio}
\begin{tabular}{@{}ll*{12}{r}@{}}
\toprule
&  & \multicolumn{4}{c}{scifact} & \multicolumn{4}{c}{trec-covid} & \multicolumn{4}{c}{touche2020} \\
\cmidrule(lr){3-6} \cmidrule(lr){7-10} \cmidrule(l){11-14}
\# & Methods & NDCG@10 & \#comp & \#inf & Lat. & NDCG@10 & \#comp & \#inf & Lat. & NDCG@10 & \#comp & \#inf & Lat. \\
\midrule
 & BM25 & 0.679 & -- & -- & -- & 0.595 & -- & -- & -- & 0.442 & -- & -- & -- \\
\midrule
\multirow{11}{*}{\rotatebox[origin=c]{90}{Flan-t5-large}}  
 & heapsort 
   & 0.675 & 222.4 & 222.4 & 26.9
   & 0.753 & 241.0 & 241.0 & 28.0
   & 0.332 & 221.0 & 221.0 & 26.0\\

 & quicksort (original, b=2) & 0.579 & 211.4 & 109.5 & 25.3 & 0.752 & 245.3 & 126.9 & 26.9 & 0.268 & 273.6 & 141.0 & 34.9 \\

 & quicksort (original, b=128) & 0.579 & 211.4 & 12.0 & 4.3 & 0.752 & 245.3 & 13.6 & 5.2 & 0.268 & 273.6 & 13.9 & 5.6 \\

 & quicksort (random, b=2) & 0.596 & 224.7 & 116.2 & -- & 0.759 & 243.8 & 126.0 & -- & 0.270 & 275.2 & 142.0 & -- \\

 & quicksort (random, b=128) & 0.611 & 218.5 & 12.3 & -- & 0.755 & 243.7 & 13.6 & -- & 0.256 & 273.0 & 13.4 & --\\

 & quicksort (middle, b=2) & 0.597 & 211.2 & 109.4 & -- & 0.763 & 235.5 & 121.8 & -- & 0.269 & 253.2 & 130.8 & -- \\

 & quicksort (middle, b=128) & 0.597 & 211.2 & 11.8 & -- &\textbf{ 0.763} & 235.5 & 13.0 & -- & 0.269 & 253.2 & 13.4 & -- \\

 & quicksort (median of three, b=2) & 0.637 & 237.1 & 107.6 & -- & \underline{0.763} & 256.0 & 115.1 & -- & 0.274 & 289.4 & 131.6 & --\\

 & quicksort (median of three, b=128) & 0.637 & 237.1 & 11.4 & -- & 0.763 & 256.0 & 12.9 & -- & 0.274 & 289.4 & 13.0 & -- \\
 & bubblesort (classic) 
   & \underline{0.692} & 805.7 & 805.7 & --
   & 0.718 & 890.1 & 890.1 & --
   & \underline{0.447} & 845.4 & 845.4 & -- \\
 & bubblesort (cached) 
   & \textbf{0.692} & 805.7 & 284.9 & --
   & 0.718 & 890.1 & 437.9 & --
   & \textbf{0.447} & 845.4 & 332.8 & --\\
\midrule
\multirow{11}{*}{\rotatebox[origin=c]{90}{Flan-t5-xl}} 
 & heapsort 
   & 0.710 & 197.5 & 197.5 & --
   & \textbf{0.783} & 249.5 & 249.5 & --
   & 0.284 & 244.3 & 244.3 & -- \\

 & quicksort (original, b=2) & 0.634 & 206.9 & 107.2 & -- & 0.761 & 225.4 & 116.6 & -- & 0.261 & 234.7 & 121.3 & --\\

 & quicksort (original, b=128) & 0.634 & 206.9 & 11.6 & -- & 0.761 & 225.4 & 12.5 & -- & 0.261 & 234.7 & 12.6 & -- \\

 & quicksort (random, b=2) & 0.646 & 219.0 & 113.3 & -- & 0.777 & 243.4 & 125.6 & -- & 0.285 & 216.3 & 112.1 & -- \\

 & quicksort (random, b=128) & 0.639 & 211.8 & 12.0 & -- & 0.772 & 222.4 & 12.8 & -- & 0.265 & 236.7 & 12.9 & -- \\

 & quicksort (middle, b=2) & 0.642 & 200.4 & 103.9 & -- & 0.777 & 239.4 & 123.6 & -- & 0.277 & 214.7 & 111.0 & -- \\

 & quicksort (middle, b=128) & 0.642 & 200.4 & 11.5 & -- & \underline{0.777} & 239.4 & 12.4 & -- & 0.277 & 214.7 & 11.9 & --\\

 & quicksort (median of three, b=2) & 0.663 & 235.1 & 106.7 & -- & 0.775 & 250.3 & 113.3 & -- & 0.283 & 232.0 & 105.1 & -- \\

 & quicksort (median of three, b=128) & 0.663 & 235.1 & 11.4 & -- & 0.775 & 250.3 & 12.4 & -- & 0.283 & 232.0 & 11.4 & -- \\

 & bubblesort (classic) 
   & \underline{0.713} & 581.9 & 581.9 & --
   & 0.748 & 874.5 & 874.5 & --
   & \underline{0.428} & 869.4 & 869.4 & --\\
 & bubblesort (cached) 
   & \textbf{0.713} & 581.9 & 217.9 & --
   & 0.748 & 874.5 & 510.9 & --
   & \textbf{0.428} & 869.4 & 467.7 & --\\
\bottomrule
\end{tabular}
\end{adjustbox}
\caption{Comparison of different methods across \textbf{SciFact}, \textbf{TREC-COVID}, and \textbf{Touche2020} datasets.}
\label{tab:dataset-comparison-2}
\end{table*}

%% file: table_trec.tex
\begin{table*}[!ht]
\footnotesize
\centering
\setlength{\tabcolsep}{2.5pt}
\begin{tabular}{@{}llrrrrrr@{}}
\toprule
\multicolumn{2}{c}{} & \multicolumn{3}{c}{TREC DL 2019} & \multicolumn{3}{c}{TREC DL 2020} \\
\cmidrule(lr){3-5} \cmidrule(lr){6-8}
\# & Methods & NDCG@10 & \#Comparisons & \#Inferences & NDCG@10 & \#Comparisons & \#Inferences \\
\midrule
 & BM25 & 0.510 & -- & -- & 0.479 & -- & -- \\
\midrule
\multirow{7}{*}{\rotatebox[origin=c]{90}{Flan-t5-large}} 
 & heapsort                      & 0.650     & 230.9 & 230.9  & \textbf{0.626}  & 226.5 & 226.5 \\
 & quicksort (original, b=2) & 0.637 & 249.0 & 128.8 & 0.588 & 237.1 & 122.7 \\
 & quicksort (original, b=128) & 0.637 & 249.0 & 14.1 & 0.588 & 237.1 & 13.5 \\
 & quicksort (random, b=2) & 0.639 & 236.7 & 122.3 & 0.587 & 236.7 & 122.4 \\
 & quicksort (random, b=128) & 0.650 & 260.5 & 13.7 & 0.580 & 240.0 & 12.8 \\
 & quicksort (middle, b=2) & 0.650 & 231.1 & 119.6 & 0.594 & 235.5 & 121.8 \\
 & quicksort (middle, b=128) & \underline{0.650} & 231.1 & 13.5 & 0.594 & 235.5 & 13.0 \\
 & quicksort (median of three, b=2) & 0.650 & 276.0 & 124.8 & 0.600 & 259.5 & 117.0 \\
 & quicksort (median of three, b=128) & \textbf{0.650} & 276.0 & 13.2 & \underline{0.600} & 259.5 & 12.8 \\
 & bubblesort (classic)          & 0.634     & 843.7 & 843.7  & 0.586           & 777.2 & 777.2 \\
 & bubblesort (cached)           & 0.634     & 843.7 & 388.3  & 0.586           & 777.2 & 357.1 \\
\midrule
\multirow{11}{*}{\rotatebox[origin=c]{90}{Flan-t5-xl}} 
 & heapsort                      & \textbf{0.706}  & 242.0 & 242.0  & \textbf{0.689}  & 244.9 & 244.9 \\
 & quicksort (original, b=2) & 0.697 & 266.6 & 137.5 & 0.672 & 250.6 & 129.3 \\
 & quicksort (original, b=128) & 0.697 & 266.6 & 13.9 & 0.672 & 250.6 & 12.9 \\
 & quicksort (random, b=2) & 0.694 & 232.3 & 120.2 & 0.676 & 239.0 & 123.5 \\
 & quicksort (random, b=128) & 0.697 & 257.3 & 13.2 & 0.676 & 237.6 & 12.7 \\
 & quicksort (middle, b=2) & 0.703 & 230.5 & 119.4 & 0.668 & 232.9 & 120.5 \\
 & quicksort (middle, b=128) & \underline{0.703} & 230.5 & 13.1 & 0.668 & 232.9 & 12.6 \\
 & quicksort (median of three, b=2) & 0.696 & 243.5 & 110.5 & 0.682 & 239.6 & 108.9 \\
 & quicksort (median of three, b=128) & 0.696 & 243.5 & 11.9 & \underline{0.682} & 239.6 & 11.8 \\
 & bubblesort (classic)          & 0.684     & 887.8 & 887.8  & 0.670           & 869.5 & 869.5 \\
 & bubblesort (cached)           & 0.684     & 887.8 & 544.9  & 0.670           & 869.5 & 542.6 \\
\midrule
\multirow{11}{*}{\rotatebox[origin=c]{90}{Flan-t5-xxl}} 
 & heapsort                      & \textbf{0.702}     & 238.9 & 238.9  & 0.688  & 239.4 & 239.4 \\
 & quicksort (original, b=2) & 0.677 & 265.9 & 137.0 & 0.680 & 234.7 & 121.3 \\
 & quicksort (original, b=128) & 0.677 & 265.9 & 13.6 & 0.680 & 234.7 & 12.7 \\
 & quicksort (random, b=2) & \underline{0.691} & 239.4 & 124.0 & 0.678 & 228.3 & 117.9 \\
 & quicksort (random, b=128) & 0.685 & 244.7 & 12.8 & 0.674 & 227.4 & 12.3 \\
 & quicksort (middle, b=2) & 0.688 & 226.3 & 117.0 & 0.677 & 229.2 & 118.5 \\
 & quicksort (middle, b=128) & 0.688 & 226.3 & 12.3 & 0.677 & 229.2 & 12.3 \\
 & quicksort (median of three, b=2) & 0.686 & 254.8 & 116.3 & \underline{0.688} & 230.4 & 104.7 \\
 & quicksort (median of three, b=128) & 0.686 & 254.8 & 11.9 & \textbf{0.688} & 230.4 & 11.4 \\
 & bubblesort (classic)          & 0.679     & 866.2 & 866.2  & 0.680           & 827.1 & 827.1 \\
 & bubblesort (cached)           & 0.679     & 866.2 & 532.1  & 0.680           & 827.1 & 465.0 \\
\midrule
\multirow{11}{*}{\rotatebox[origin=c]{90}{\tiny{Meta-Llama-3-8B-Instruct}}} 
 & heapsort                      & \underline{0.662}     & 235.0 & 235.0  & \textbf{0.615}  & 231.9 & 231.9 \\
 & quicksort (original, b=2) & 0.645 & 266.5 & 137.4 & 0.576 & 235.5 & 121.8 \\
 & quicksort (original, b=128) & 0.645 & 266.5 & 13.5 & 0.576 & 235.5 & 12.9 \\
 & quicksort (random, b=2) & \textbf{0.663} & 231.3 & 119.8 & 0.580 & 231.7 & 119.8 \\
 & quicksort (random, b=128) & 0.660 & 219.0 & 12.8 & 0.585 & 232.9 & 12.8 \\
 & quicksort (middle, b=2) & 0.640 & 220.9 & 114.4 & 0.564 & 228.1 & 118.0 \\
 & quicksort (middle, b=128) & 0.640 & 220.9 & 12.6 & 0.564 & 228.1 & 12.6 \\
 & quicksort (median of three, b=2) & 0.650 & 236.0 & 106.3 & 0.594 & 244.3 & 110.1 \\
 & quicksort (median of three, b=128) & 0.650 & 236.0 & 12.0 & 0.594 & 244.3 & 12.3 \\
 & bubblesort (classic)          & 0.641 & 822.5 & 822.5    & 0.600 & 797.6 & 797.6    \\
 & bubblesort (cached)           & 0.641     & 822.5 & 389.4  & \underline{0.600}           & 797.6 & 365.9 \\
\midrule
\multirow{11}{*}{\rotatebox[origin=c]{90}{\tiny{Mistral-7B-Instruct-v0.1}}} 
 & heapsort                      & 0.559     & 200.3 & 200.3  & 0.513           & 190.1 & 190.1 \\
 & quicksort (original, b=2) & 0.578 & 278.9 & 143.7 & 0.529 & 276.2 & 142.2 \\
 & quicksort (original, b=128) & 0.578 & 278.9 & 14.0 & 0.529 & 276.2 & 13.4 \\
 & quicksort (random, b=2) & 0.593 & 293.8 & 150.9 & 0.511 & 271.5 & 139.8 \\
 & quicksort (random, b=128) & 0.573 & 279.7 & 12.9 & 0.524 & 263.9 & 13.2 \\
 & quicksort (middle, b=2) & 0.595 & 257.3 & 132.7 & 0.531 & 249.3 & 128.6 \\
 & quicksort (middle, b=128) & 0.595 & 257.3 & 13.5 & 0.531 & 249.3 & 12.9 \\
 & quicksort (median of three, b=2) & \underline{0.612} & 292.1 & 132.5 & 0.538 & 292.5 & 133.0 \\
 & quicksort (median of three, b=128) & \textbf{0.612} & 292.1 & 13.1 & 0.538 & 292.5 & 12.9 \\
 & bubblesort (classic)          & 0.587 & 631.0 & 631.0  & \underline{0.539} & 578.7 & 578.7 \\
 & bubblesort (cached)           & 0.587     & 631.0 & 250.5  & \textbf{0.539}  & 578.7 & 223.4 \\
\bottomrule
\end{tabular}
\caption{Comparison of methods for \textbf{TREC DL 2019} and \textbf{TREC DL 2020}.}
\end{table*}